# Taiyi: A Bilingual Fine-Tuned Large Language Model for Diverse Biomedical Tasks

Ling Luo*, Jinzhong Ning, Yingwen Zhao, Zhijun Wang, Zeyuan Ding, Peng Chen, Weiru Fu, Qinyu Han, Guangtao Xu, Yunzhi Qiu, Dinghao Pan, Jiru Li, Hao Li, Wenduo Feng, Senbo Tu, Yuqi Liu, Zhihao Yang, Jian Wang, Yuanyuan Sun, Hongfei Lin

School of Computer Science and Technology, Dalian University of Technology, Dalian 116024, China
*Corresponding author: lingluo@dlut.edu.cn

## Abstract

**Objective**: Most existing fine-tuned biomedical large language models (LLMs) focus on enhancing performance in monolingual biomedical question answering and conversation tasks. To investigate the effectiveness of the fine-tuned LLMs on diverse biomedical NLP tasks in different languages, We present Taiyi, a bilingual fine-tuned LLM for diverse biomedical tasks.

**Materials and Methods**: We first curated a comprehensive collection of 140 existing biomedical text mining datasets (102 English and 38 Chinese datasets) across over 10 task types. Subsequently, a two-stage strategy is proposed for supervised fine-tuning to optimize the model performance across varied tasks.

**Results**: Experimental results on 13 test sets covering named entity recognition, relation extraction, text classification, question answering tasks demonstrate that Taiyi achieves superior performance compared to general LLMs. The case study involving additional biomedical NLP tasks further shows Taiyi's considerable potential for bilingual biomedical multi-tasking.

**Conclusion**: Leveraging rich high-quality biomedical corpora and developing effective fine-tuning strategies can significantly improve the performance of LLMs within the biomedical domain. Taiyi shows the bilingual multi-tasking capability through supervised fine-tuning. However, those tasks such as information extraction that are not generation tasks in nature remain challenging for LLM-based generative approaches, and they still underperform the conventional discriminative approaches of smaller language models.

## INTRODUCTION

Recently, the release of ChatGPT [1] and the subsequent launch of GPT-4 [2] have received widespread attention around the world, which promotes the development of large language models (LLMs) that have billions of parameters and trained with hundreds of billions of tokens. These LLMs (such as GPT-4, PaLM [3], LLaMA [4], and GLM [5]) have shown promising results and achieved new state-of-the-art performance in various natural language processing (NLP) tasks. Compared with previous pre-trained language models (such as BERT [6]), LLMs exhibit enhanced text generation and comprehension abilities. Moreover, their few-shot/zero-shot learning and generalization abilities address unseen or complicated tasks. Due to the advantages, LLMs have been explored to apply in various fields, such as law [7], education [8], finance [9], and biomedicine and health [10].

LLMs in the biomedical and healthcare domain are essential and potential for handling the scale and complexity of biomedical data, as well as for providing more personalized and empathetic medical care, ultimately advancing the quality and effectiveness of healthcare practices. However, biomedical text exhibits some distinct features compared to text in the general domain, such as complex terminologies, ambiguous abbreviations, more complex syntax, and less common vocabulary. Owing to these domain characteristics, most existing LLMs trained on general domain data encounter difficulties when they are applied to the biomedical domain [11]. Therefore, some domain-special LLMs have been developed to facilitate the development of NLP methods for biomedical applications. For example, Google's team recently developed Med-PaLM2 [12], a fine-tuned LLM using medical data based on PaLM [3], which has achieved a high performance on the medical QA datasets with an accuracy of over 80%. Apart from the non-open biomedical LLMs, there have been some efforts to develop open-source biomedical LLMs [13-20]. Most of the models are derived from some open-source general LLMs (such as LLaMA [4], BLOOMZ [21], and GLM [5]) and are fine-tuned with monolingual (e.g., English or Chinese) question answering (QA) and conversation data. They are primarily directed toward enhancing performance in biomedical QA and conversation-oriented tasks. However, the efficacy and suitability of the LLMs on diverse biomedical NLP tasks in different languages remains unclear and warrants further investigation.

To address these problems, we present Taiyi, a bilingual (English and Chinese) fine-tuned large language model for diverse biomedical tasks. First, a comprehensive collection including 102 English and 38 Chinese datasets is assembled, covering over 10 biomedical task types. To facilitate task-specific requirements and enable consistent formatting across all datasets, standardized data schemas are designed and universally applied during dataset curation. Then, in the supervised fine-tuning (SFT) phase, we propose a two-stage fine-tuning strategy. In contrast to the simple single-stage fine-tuning, this strategy significantly optimizes model performance across a diversity of tasks. Finally, the evaluation of Taiyi is conducted on 13 Biomedical NLP test sets. Experimental results demonstrate the promising potential of Taiyi in bilingual multi-task learning. It achieves superior performance on biomedicine-specific tasks compared to general LLMs. However, Taiyi has an improvement room for current state-of-the-art models specialized for individual tasks.

## RELATED WORK

Recent advancements in LLMs, such as GPT-4 [2] and PaLM [3], have attracted considerable attention due to instruction-following and producing human-like responses. Subsequently, the researchers attempt to duplicate the GPT series to develop open-source foundation models like LLaMA [4], Bloom

[22], Falcon [23], GLM [5] and Qwen [24]. These LLMs present strong performances on various NLP tasks, including zero- and few-shot learning scenarios. The promising capabilities of LLMs have sparked interest and potential applications in various fields, particularly in the biomedical domain.

When applied to specific domains like biomedicine, large-scale models often perform sub-optimal. To improve the performance of the models in biomedical tasks, there have been some efforts in training LLMs specifically for the biomedical domain. For example, MedAlpaca [14] builds upon medical data to fine-tune Stanford Alpaca for applications related to medical question-answering and dialogue. ChatDoctor [19] is designed to simulate a conversation between a doctor and a patient, fine-tuning LLaMA with medical literature. Additionally, Med-PaLM [25] has shown promising performance on the MedQA exam based on clinical corpora and human feedback. Meanwhile, aiming at the Chinese medical domain, Chinese LLMs such as BenTsao [16], DoctorGLM [17], and HuatuoGPT [20], are developed on the Chinese medical dialogue data. And more recently Zhongjing [18] and ChiMed-GPT [26] adopted full pipeline training from pre-training, SFT, to reinforcement learning with human feedback. The overview of existing LLMs in the biomedical domain can be found in Supplementary Table S1. Most existing open-source LLMs focus on fine-tuning with monolingual QA and conversation data. Different from the above monolingual LLMs, our Taiyi embarks on the bilingual (English and Chines) biomedical large language model, aiming to explore the capabilities of large models in handling a variety of bilingual NLP tasks in the biomedical field.

## MATERIALS AND METHODS

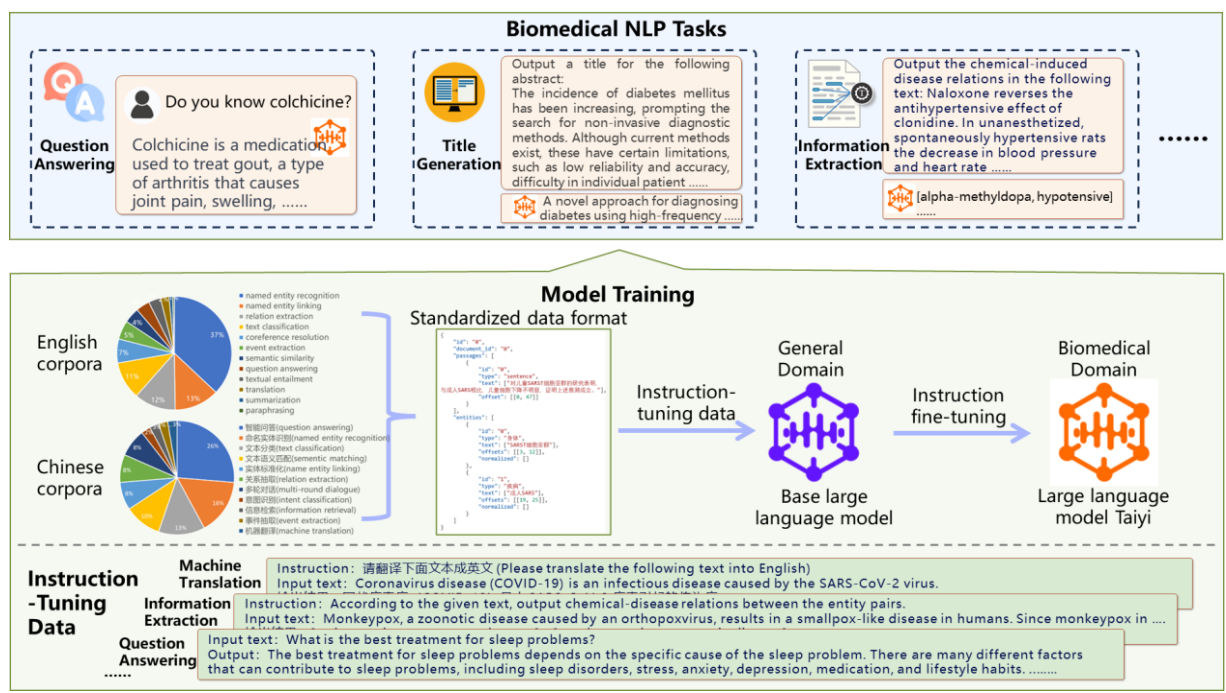

**Figure 1**. Overview of the Taiyi framework. Note that the English in brackets is not part of the input and output; it is the translation of the Chinese.

Figure 1 shows the overview of our framework. We first collected and curated a set of existing bilingual (Chinese and English) biomedical natural language processing (BioNLP) corpora. Then these corpora were converted to the instruction data used to fine-tune the general LLM. In the supervised fine-tuning phase, we propose a two-stage supervised instruction fine-tuning strategy. Finally, our Taiyi model can be applied to various BioNLP tasks and it is evaluated on 13 test sets covering 4 BioNLP task types.

**Training data**

**Dataset collection**

This study aims to explore the capabilities of LLMs in handling a variety of bilingual BioNLP tasks by supervised fine-tuning. Therefore, we focus on manually annotated English and Chinese biomedical corpora. To fully utilize available BioNLP resources, we make efforts to aggregate comprehensive sets of open-source datasets in both English and Chinese. The data collection is primarily from two sources: existing English/Chinese BioNLP shared task datasets and the training data used for existing biomedical LLMs.

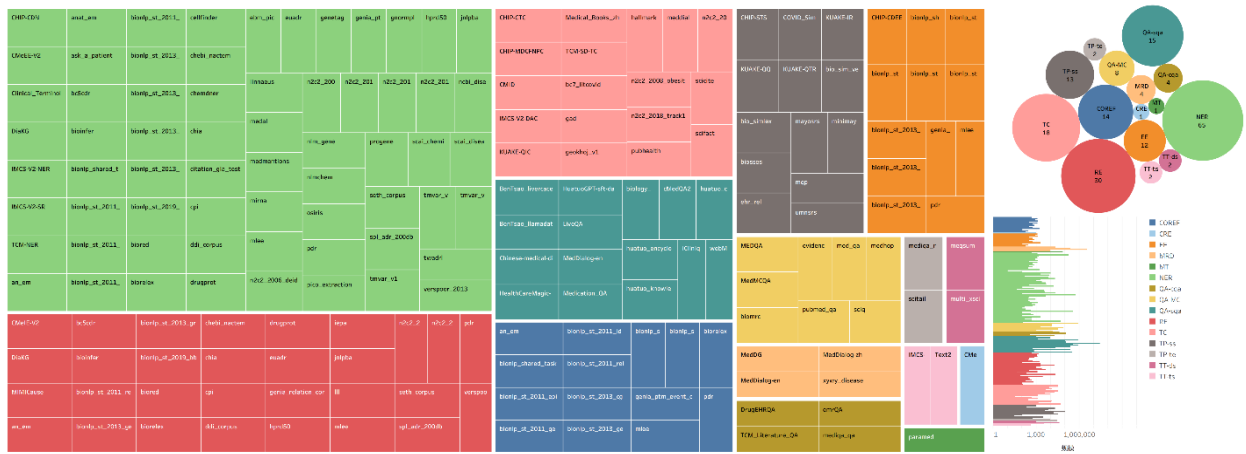

**Figure 2**. Overview of the dataset collection. In the tree map (left), the entire graph is represented by a large rectangle, representing the 140 bilingual open-source datasets collected. This large rectangle is divided into smaller rectangles of various colors, with each small rectangle representing a specific BioNLP task and listing the names of all datasets related to that task. In the scatter plot (top right), each data point represents a BioNLP task, with its size determined by the number of datasets associated with that task. The bar chart (bottom right) presents the scale of related datasets for different tasks, further illustrating the number of datasets for each task.

Our data collection significantly benefited from two major previous efforts in aggregating biomedical text mining datasets – BigBio [27] and CBLUE [28]. The BigBio aggregates a large collection of English BioNLP datasets, while the CBLUE dataset assembles a wide range of Chinese biomedical natural language understanding datasets. In addition, we also collected some other relevant BioNLP datasets that are not included in BioBio and CBLUE. Finally, we have successfully assembled a total of 140 biomedical datasets. All datasets are categorized into 15 different BioNLP task types as shown in Figure 2: Named Entity Recognition/Normalization (NER/NEN), Relation Extraction (RE), Causal Relation Extraction (CRE), Event Extraction (EE), Coreference Resolution (COREF), Text Classification (TC), Question Answering-Multiple Choice (QA-mc), Question Answering-Simple Answer Questions (QA-sqa), Question Answering-Context-based Answer Questions (QA-cqa), Multi-Round Dialogue (MRD), Machine Translation (MT), Text Pairs-Semantic Similarity (TP-ss), Text Pairs-Textual Entailment (TP-te), Text to Text/Struct-Document Summarization (TT-ds) and Text/Struct-Text to Struct (TT-ts). Among these datasets, there are 38 Chinese datasets covering 10 different BioNLP tasks, and 102 English datasets spanning 12 BioNLP tasks.

For each dataset, we collated key metadata including task types, data size, task descriptions, and the links of the dataset and paper. This metadata facilitates full understanding and proper usage of each corpus. The collection of bilingual datasets across diverse biomedical language processing tasks facilitates the comprehensive evaluation and development of multilingual BioNLP models. Details can be accessed at the following link: https://github.com/DUTIR-BioNLP/Taiyi-LLM/blob/main/data_file/dataset_inf.md.

**Task schema harmonization**

In the original data set, even for the same BioNLP task, there are many different data formats due to different sources and developers. For example, the annotation formats include BioC, CoNLL, and PubTator formats across NER datasets. Such inconsistencies introduce challenges for developing systems that can leverage diverse annotated datasets. Therefore, establishing a universal data format enables interoperability and is beneficial for cross-dataset integration. To be consistent with previous efforts, we extended the schema from BioBio schema to support all tasks in our collection. The complete unified task schema can be found at https://github.com/DUTIR-BioNLP/Taiyi-LLM/blob/main/data_file/Task_schemas_en.md.

**Instruction-tuning data construction**

To enable the model to understand task instructions for performing multi-tasking, we constructed the instruction data for fine-tuning, which covers the tasks described in the above section. Moreover, we also added MOSS data (i.e., Moss-003-sft-data) [29] into the training data to enhance the general conversation and harmlessness ability of Taiyi. Moss-003-sft-data is well-aligned with the real-world distribution of user intents, covering finer-grained categories and more diverse harmlessness-related data. Previous studies [30 31] have demonstrated that the quality of the training data plays a critical role in the performance of downstream tasks. Inferior quality data adversely affects the performance of the LLMs. Therefore, we manually analyzed the datasets, then the high-quality datasets were selected. We filtered duplicated training data, and those overlapped documents in the training set if the documents also exist in the test set to accurately evaluate the model performance. The statistics of the final data used for instruction-tuning is presented in Table 1.

Table 1. Statistics of our final instruction-tuning data

| Task Type | English data size | Chinese data size |
|---|---|---|
| Named Entity Recognition | 28,603 | 44,667 |
| Relation Extraction | 17,279 | 26,606 |
| Event Extraction | 2,022 | 2,992 |
| Text Classification | 40,339 | 37,624 |
| Text Pair Task | 11,237 | 45,548 |
| Machine Translation | 74,113 | |
| Biomedical Question Answering | 57,962 | 129,562 |
| Biomedical Multi-Round Dialogue | 10,000 | 16,391 |
| General Dialogue Data | 560,000 | |
| Other Additional Tasks | 9,370 | |
| *Total* | 1,114,315 | |

To construct the instruction-tuning data, we design instructional templates for each task. For the QA and dialogue tasks, original questions are used as the model input and answers are used as the output. For other tasks, approximately 15 instruction templates were manually created for each task in English and Chinese, respectively. Some examples of the instruction data can be found in Supplementary Table S2.

Furthermore, some complex NER and RE tasks are divided into multiple subtasks to reduce difficulty and increase diversity. For instance, the BC5CDR [32] task of chemical and disease entity recognition is separated into the chemical entity recognition and disease entity recognition subtasks. The original complex task is retrained while these additional subtasks are added to the instruction-tuning data.

**Model training**

**Base model**

Recently, Alibaba Cloud's pre-trained LLM Qwen series [24] have been developed and shown promising results in various NLP tasks. We chose the Qwen-7B-base version for supervised instruction fine-tuning. Qwen-7B is a Transformer-based pre-trained language model that obtains generalized language understanding capabilities through self-supervised learning on large-scale high-quality multilingual pretraining corpora. Compared to models like BERT, the training data coverage of Qwen-7B is more extensive, including web texts, academic books, code, and other resources. It has approximately 7 billion parameters and a vocabulary size of around 150,000.

In public leaderboards[1,2] of downstream English and Chinese tasks, Qwen-7B significantly outperforms models of similar size and even surpasses larger models on some tasks. This demonstrates Qwen-7B's strong language understanding and transfer learning abilities. The considerations for choosing Qwen-7B as our pre-trained model are as follows: 1) The model size is moderate, with high training efficiency that meets our computational resource constraints; 2) The training data coverage is extensive, encompassing the common language, domain-specific language, and code data required for our cross-domain task; 3) It achieves strong performance on public benchmarks, with sufficient language understanding capabilities to provide a solid foundation for our downstream task.

---

[1] https://huggingface.co/spaces/HuggingFaceH4/open_llm_leaderboard;
[2] https://cevalbenchmark.com/static/leaderboard.html

**Two-stage Supervised Instruction Fine-tuning**

During the SFT stage, we performed fine-tuning on our curated dataset of millions of examples. Since the instruction-tuning data involves dozens of datasets across over ten biomedical NLP tasks in both Chinese and English, performing fine-tuning in a simple single stage would result in task interference, preventing convergence on some of the more challenging NLP tasks. Therefore, we propose a two-stage supervised instruction fine-tuning strategy shown in Figure 3.

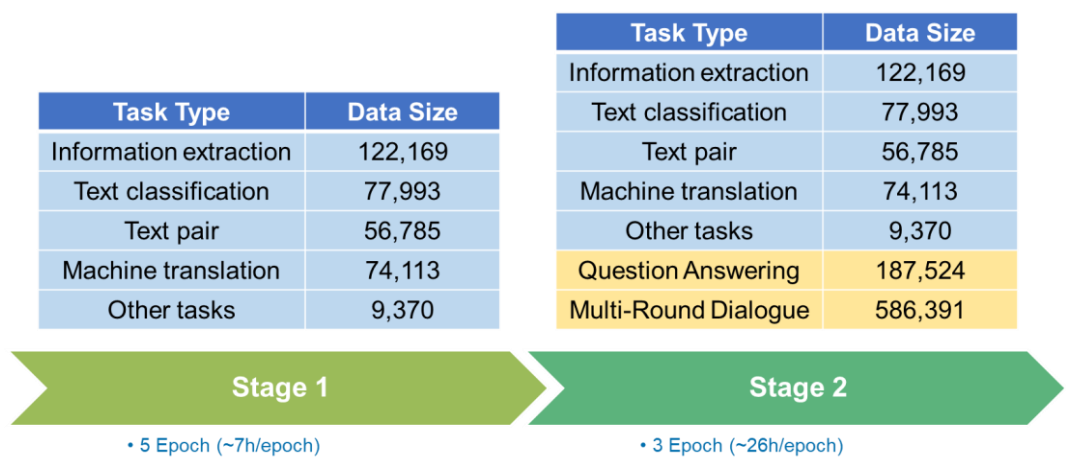

**Figure 3**. The two-stage training process of Taiyi. Tasks in Type1 are in the blue background, and task2 in Type2 are in the yellow background.

According to the task type and the size of the task dataset, we first manually categorized all the tasks into two types of the tasks: Type1 and Type2. In Type1, most tasks are not generation tasks in nature (e.g., NER is a sequence labeling task in nature rather than a generation task), or the size of the task dataset is related small. These tasks include information extraction, text classification, text pair tasks, machine translation and other additional tasks. In Type2, there are QA and dialogue tasks including biomedical QA, biomedical multi-round dialogue, and general dialogue tasks. These tasks are generation tasks in nature. In the first stage of model training, we first performed supervised instruction fine-tuning on the data from the Type1 tasks (around 340,000 instances). The best checkpoint was selected through a combination of human evaluation and automated metrics on the development sets for the second stage of training. In the second training stage, all the training data from the first stage are used as retrospective data, and all data in Type2 are mixed for incremental training.

We used 8 A40 GPUs for SFT. The model was trained with 5 epochs (~7 hours per epoch) in stage 1, and 3 epochs (~26 hours per epoch) in stage 2. To improve model training efficiency, we chose Qlora (Dettmers, et al., 2023), an efficient tuning algorithm, to conduct supervised instruction fine-tuning. The main hyperparameters for the training process are set as follows: Batch size per GPU of 12, learning rate of 0.0002, warmup ratio of 0.1, max length of 1024, lora rank of 64, lora alpha of 16, and lora dropout of 0.05.

## RESULTS

### Evaluation tasks, datasets and metrics

To investigate the capability of the Taiyi model for various bilingual BioNLP tasks, we selected four task types (i.e., NER, RE, TC, multiple-choice QA) as the metrics evaluation and other tasks for the case study. The statistics of the test sets are shown in Supplementary Table S3. Further details for each evaluation task are provided below.

**Biomedical Named Entity Recognition (NER)**. Biomedical NER aims to identify predefined biomedical entities from text, such as diseases, drugs, genes, and proteins. Six biomedical NER datasets (i.e., BC5CDR-Chemcial [32], BC5CDR-disease [32], CHEMDNER [33], NCBI-Disease [34], BioRED [35] and CMeEE-dev [28]) are used for our NER evaluation. The exact match (the predicted entity text and entity type without span) micro F1-score was used as the evaluation metrics.

**Biomedical Relation Extraction (RE)**. Biomedical RE aims to automatically extract predefined relations or associations between biomedical entities from text. In this task, we focus on extracting the entity relation triple (head entity, tail entity, relation type). We selected one English dataset BC5CDR and one Chinese dataset CMeEE-dev [28] for the RE evaluation. The micro F1-score was used as the RE metrics.

**Biomedical Text Classification (TC)**. Biomedical TC aims to automatically categorize texts into predefined biomedical class labels. Two manually annotated English multi-label document classification datasets (BC7LitCovid [36] and HoC [37]) and one Chinese dataset (KUAKE_QIC [28]) are used for the TC evaluation. The micro F1-score was used as the metrics.

**Biomedical Multiple-choice Question Answering (QA-mc)**. Biomedical multiple-choice Question Answering task aims to answer multiple-choice questions related to biomedicine and healthcare. PubMedQA [38] and MedQA-USMLE (4-option) [39] are used to evaluate the model performance in English. MedQA-MCMLE (4-option) [39] is used to evaluate the model performance in Chinese. We used the accuracy as the metrics for this task.

In addition to the four primary evaluation tasks, the capabilities of the Taiyi model were also shown through examples on several supplementary biomedical NLP tasks, such as medical Report Generation, Biomedical Event Extraction, Biomedical Machine Translation, Biomedical Title Generation, Biomedical Text Semantic Similarity, Biomedical Question Answering and Chat. Details can be found in Supplementary Examples of Taiyi output on the supplementary biomedical NLP tasks.

**Effectiveness of the two-stage fine-tuning strategy**

In this experiment, we tested the effect of our two-stage fine-tuning strategy on the four tasks. For comparison, we simply combined all data of the four tasks to finetune the base model as the baseline. As the cost of LLMs is high, we randomly sampled 200 instances from the test set per dataset. The performances of the models using the simple one-stage and the two-stage strategies on the tasks are shown in Figure 4.

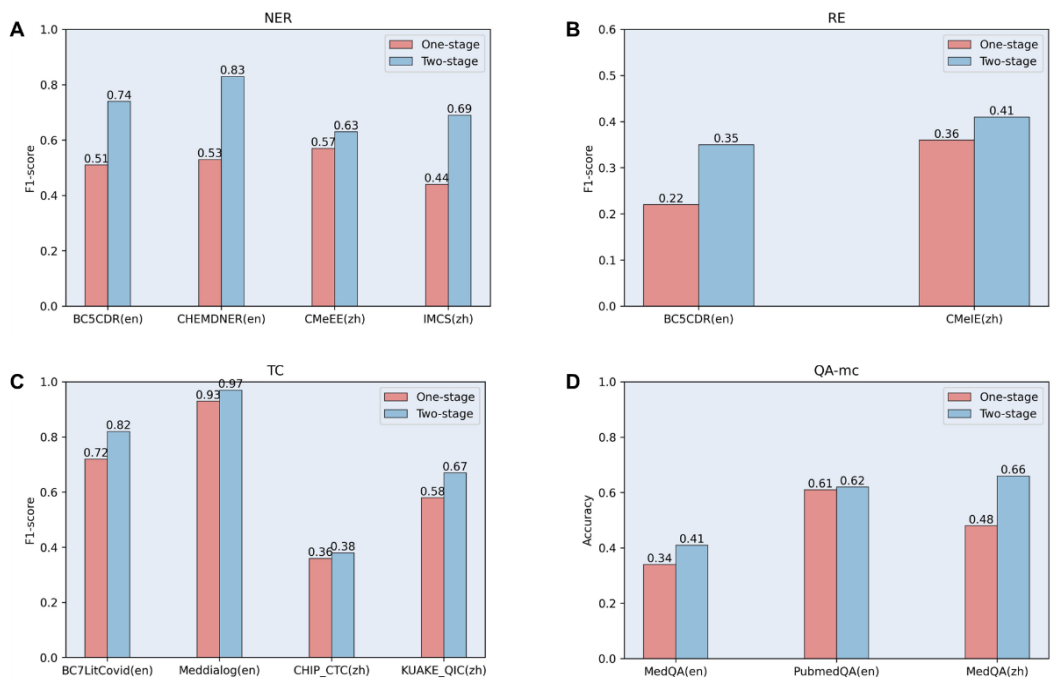

**Figure 4**. Performances of models using different fine-tuning strategies. One-stage denotes the model fine-tuned by combining all task datasets. Two-stage denotes the model fine-tuned by our two-stage strategy. (A) the results on the NER tasks. (B) the results on the RE tasks. (C) the results on the TC tasks. (D) the results on the QA-mc tasks.

When we simply combined all task datasets as a one-stage strategy to fine-tune the LLM, the results show poor performances on all tasks. The main reason may be the challenges of converging different tasks with varying levels of difficulty and dataset sizes. Therefore, we first fine-tuned the model on those Type1 tasks described in the section of Two-stage Supervised Instruction Fine-tuning independently, then combined all data to continue fine-tuning the model in the second stage. The results show that our two-stage strategy outperforms the one-stage strategy on all English and Chinese tasks and achieves significant average improvement (~10% in metrics). The two-stage approach allows the model to first become specialized on tasks that are not generation tasks in nature before developing more generalized capabilities across tasks in the second stage.

**Performance of Taiyi on the entire test sets**

The previous experiment demonstrates our two-stage approach achieves significant improvements on the subset of the test sets compared to the simple one-stage training strategy. In this experiment, we evaluate the performance of our Taiyi on the entire test sets of the 13 tasks covering 4 BioNLP task types. The results of ChatGPT 3.5 (GPT-3.5-Turbo) and the state-of-the-art (SOTA) methods based on supervised pretrained language models for each corpus are provided for comparison.

**Table 2**. Performance comparison with other existing methods on the 13 BioNLP tasks

| Task | Datasets | Taiyi | ChatGPT3.5 | SOTA |
|---|---|---|---|---|
| NER (Micro-F1) | BC5CDR-Chem (en) | 0.802 | 0.603 [40] | 0.933 (PubMedBERT [40]) |
| | BC5CDR-Dise (en) | 0.691 | 0.518 [40] | 0.856 (PubMedBERT [40]) |
| | CHEMDNER (en) | 0.799 | 0.365 [41] | 0.924 (BioBERT [42]) |
| | NCBI-Disease (en) | 0.731 | 0.505 [40] | 0.878 (PubMedBERT [40]) |
| | CMeEE (zh) | 0.657 | 0.470 [43] | 0.740 CBLUE-Leaderboard [28] ) |
| RE (Micro-F1) | BC5CDR (en) | 0.375 | 0.142 | 0.450 (BioGPT [44]) |
| | CMeIE (zh) | 0.432 | 0.306 [43] | 0.549 CBLUE-Leaderboard [28]) |
| TC (Micro-F1) | BC7LitCoivd (en) | 0.840 | 0.639 [45] | 0.918 (Bioformer [46]) |
| | HOC (en) | 0.800 | 0.512 [40] | 0.823 (PubMedBERT [40]) |
| | KUAKE_QIC (zh) | 0.774 | 0.485 [43] | 0.859 (CBLUE-Leaderboard [28]) |
| QA (Accuracy) | PubMedQA (en) | 0.544 | 0.765 [40] | 0.558 (PubMedBERT [40]) |
| | MedQA-USMLE (en) | 0.371 | 0.513 [47] | 0.367 (BioBERT-large [39]) |
| | MedQA-MCMLE (zh) | 0.648 | 0.582 [47] | 0.701 (RoBERTA-large [39]) |
| ALL | AVE | 0.651 | 0.493 | 0.735 |

Note: For the results of ChatGPT3.5 and SOTA methods, we provided previous results on the published papers. Since the published result of ChatGPT 3.5 is not available for the relation extraction on the BC5CDR test set, we obtained the result of the ChatGPT using our same prompt via the OpenAI API. For the CMeEE, CMeIE and KUAKE_QIC datasets, the results of Taiyi are reported on the development set since the gold standard test sets have not been released. Although they may not be directly compared, we still list them for reference.

As shown in Table 2, our Taiyi outperforms ChatGPT3.5 on 11 out of 13 datasets, except for two English QA datasets. It shows that the fine-tuned LLM on domain-specific instruction data can improve performance on in-domain downstream tasks. However, the lower English QA results suggest that the rich domain knowledge may be learned from the pre-training stage of the LLMs, which is difficult to compensate via instruction fine-tuning. Compared with SOTA pre-trained language models, the Taiyi model achieves comparable results on the QA tasks. However, the performance of Taiyi still falls by a margin (average ~ 9% in the metrics) for the NER, RE, and TC tasks, where conventional discriminative methods outperform generative methods.

**Performance of Taiyi on the new task**

To investigate whether the Taiyi model can be applied to support the new task that is not previously seen in our fine-tuning data, we hold out a recently published biomedical corpus BioRED for evaluation. BioRED is a biomedical relation extraction dataset with multiple entity types (e.g. gene/protein, disease, chemical) and relation pairs (e.g. gene–disease; chemical–chemical) at the document level. Here, we focus on the NER performance of Taiyi on the BioRED test set.

Table 3. Performance of the Taiyi model on the BioRED test set

| Entity Type | P | R | F1 |
|---|---|---|---|
| Chemical | 0.717 | 0.566 | 0.633 |
| Disease | 0.829 | 0.533 | 0.649 |
| Gene | 0.931 | 0.490 | 0.642 |
| Variant | 0.633 | 0.585 | 0.608 |
| Species | 0.640 | 0.286 | 0.395 |
| Celline | 0.213 | 0.565 | 0.310 |
| *All* | 0.760 | 0.507 | 0.608 |

As shown in Table 3, Taiyi obtains promising results (overall performance of 0.608 in F1-score) on the new BioRED NER task. It is significantly higher than the performance of ChatGPT (0.391 in F1-score [41]) on the test set. Among these entity types, Taiyi achieves better performance to chemical, disease, gene, and variant entities, since these entities are seen in other related datasets (e.g., BC5CDR, GnormPlus, and tmVar) during the SFT stage. Moreover, it is noteworthy that even for the unseen entity types (i.e., species and cell lines) in the SFT stage, Taiyi still understands the entity recognition instruction and extracts some species and cell line entities. These results suggest that Taiyi is highly robust and generalizable.

**Case study**

To further investigate the benefits of the in-domain instruction fine-tuning, we qualitatively analyzed model outputs on various BioNLP tasks and compared Taiyi to Qwen7B-Chat, a published LLM fine-tuned using general domain instruction data on the same base model Qwen7B. Table 4 shows answers produced by different models with the same questions on the Chinese chat. From the results, Taiyi generates more detailed and accurate responses to biomedical questions compared to the general domain model Qwen-Chat. More examples of different BioNLP tasks can be found in the Supplementary Case study.

**Table 4**. Examples of generated text on the Chinese chat task

**Input (1)**:
什么是鼠疫？(What is plague?)
**Taiyi**:
鼠疫是一种由鼠疫杆菌引起的传染病，主要通过跳蚤等媒介传播。鼠疫杆菌主要存在于啮齿类动物体内，当啮齿类动物死亡后，跳蚤等昆虫会吸食其体液，从而将鼠疫杆菌带入人体，引起鼠疫感染。鼠疫的症状包括高热、寒战、淋巴结肿大、咳嗽、呼吸急促等，严重时可导致死亡。鼠疫是一种古老的传染病，历史上曾多次引起大规模的流行病，如14世纪的黑死病，对欧洲造成了巨大的人员和经济损失。(Plague is an infectious disease caused by Yersinia pestis and is mainly spread through fleas and other vectors. Yersinia pestis mainly exists in rodents. When rodents die, fleas and other insects will suck their body fluids, thereby introducing Yersinia pestis into the human body, causing plague infection. Symptoms of plague include high fever, chills, swollen lymph nodes, cough, shortness of breath, etc., which can lead to death in severe cases. Plague is an ancient infectious disease that has caused large-scale epidemics many times in history, such as the Black Death in the 14th century, which caused huge human and economic losses to Europe.)
**Qwen-Chat**:
鼠疫是一种由鼠疫杆菌引起的急性传染病，它可以在动物之间传播，并且可以通过接触感染病动物的排泄物或携带病毒的动物及其分泌物传播给人类。该病的症状包括发热、寒战、淋巴结肿大、呼吸困难和出血等。(Plague is an acute infectious disease caused by Yersinia pestis, which can be transmitted between animals and can be transmitted to humans through contact with the excrement of infected animals or animals carrying the virus and their secretions. Symptoms of the disease include fever, chills, swollen lymph nodes, difficulty breathing and bleeding.)

**Input (2)**:
有什么预防方法吗？(Are there any prevention methods?)
**Taiyi**:
是的，预防鼠疫的方法包括以下几点：
1. 避免接触感染源：尽量避免接触啮齿类动物，如老鼠、土拨鼠等，不要在它们的巢穴附近逗留。
2. 注意个人卫生：保持手部卫生，勤洗手，避免接触有感染风险的物品，如被感染的衣物、被褥等。
3. 食用安全：避免食用被感染的动物或其制品，如鼠肉、鼠肉制品等。
4. 家居环境：保持家居环境清洁卫生，定期消毒，特别是鼠洞、鼠粪等易滋生细菌的地方。
5. 预防接种：对于有鼠疫高发区的人群，可以接种鼠疫疫苗，提高免疫力。 总之，预防鼠疫需要从多个方面入手，保持个人卫生、注意饮食安全、保持家居环境清洁卫生等都是预防鼠疫的有效方法。
(Yes, ways to prevent plague include the following:
1. Avoid contact with sources of infection: Try to avoid contact with rodents, such as mice, prairie dogs, etc., and do not stay near their nests.
2. Pay attention to personal hygiene: maintain hand hygiene, wash hands frequently, and avoid contact with items with risk of infection, such as infected clothing, bedding, etc.
3. Food safety: Avoid eating infected animals or their products, such as rat meat and rat meat products.
4. Home environment: Keep the home environment clean and sanitary, and disinfect it regularly, especially rat holes, rat droppings and other places where bacteria can easily breed.
5. Vaccination: People in areas with a high incidence of plague can be vaccinated against plague to improve their immunity. In short, the prevention of plague requires many aspects. Maintaining personal hygiene, paying attention to food safety, and keeping the home environment clean and hygienic are all effective ways to prevent plague.)
**Qwen-chat**:
预防鼠疫的最有效方法是避免接触携带鼠疫杆菌的动物和它们的排泄物，以及不触摸被污染的物品和表面。另外，应保持个人卫生，经常洗手，避免进食未煮熟的食物，以及及时接种疫苗来预防鼠疫的发生。(The most effective way to prevent plague is to avoid contact with animals that carry Yersinia pestis and their excrement, as well as not touching contaminated objects and surfaces. In addition, you should maintain personal hygiene, wash your hands frequently, avoid eating undercooked food, and get vaccinated in time to prevent the occurrence of plague.)

Note that the English in brackets is not part of the input and output; it is the translation of the Chinese.

## DISCUSSION

Due to the complexity of biomedical data and knowledge gaps across tasks, general LLMs often generate meaningless or completely unrelated responses to a given biomedicine input, or the users need to design professional prompts to improve the performance of LLMs on the biomedical tasks. To mitigate such issues, we elaborately fine-tuned the LLM on bilingual biomedical data across multiple biomedical tasks. For the LLM optimization to biomedicine, it is crucial to establish a data-centric infrastructure and fine-tuning procedure. In our study, we collected a huge amount of in-domain training datasets, filtered out high-quality subsets and optimized reasonable ratios for each dataset. Including the tasks that are not generation tasks in nature (e.g., NER and RE), all tasks are converted into generation tasks via instruction templates to train the model. The proposed two-stage SFT strategy is shown to effectively improve the model performance. Owing to these efforts, Taiyi can understand bilingual biomedical task instructions and has strong robustness and generalization capabilities on a variety of tasks.

However, Taiyi still has some common limitations of LLMs, including hallucinations, bias and fairness, lack of common sense, and deficient biomedical knowledge. For example, when we input the sentence "Please introduce the octacyclines among antibiotics", Taiyi generated the response of "Octacyclines are a group of antibiotics that are structurally similar to each other and share a common chemical structure. They are characterized by having a 1,4-dioxane ring in their core structure……". In fact, "Octacyclines" are fictitious and there are no such antibiotics. The hallucinations may potentially lead to severe medical malpractice. Moreover, Taiyi achieves better performance on the Chinese QA task but lower scores on the English QA tasks than ChatGPT 3.5 as shown in Table 2. This suggests that the rich biomedical knowledge may be learned from the pre-training stage of the LLMs, which is difficult to compensate via simple instruction fine-tuning. In practice, it is very challenging to train capable biomedical LLMs from scratch, due to the huge compute consumption and the sensitivity to data quality and training tricks. Therefore, it is useful to develop effective tuning strategies and use additional biomedical resources to inject specific knowledge. Our future work will focus on these problems, incorporating knowledge resources (e.g., biomedical knowledge database and factual information obtained by retrieval technology), improving the biomedical interpretability of the model's output, and aligning with human intentions to improve safety in the medical field.

## CONCLUSION

In this study, we collected diverse bilingual (English and Chinese) BioNLP datasets and standardized their formats. Leveraging these rich training resources and the proposed two-stage supervised fine-tuning approach, Taiyi shows considerable capability on various BioNLP tasks. Furthermore, Taiyi exhibits cross-lingual generalization across similar task scenarios while retaining general domain conversational abilities. Overall, utilizing rich high-quality biomedical corpora and designing effective fine-tuning strategies can substantially enhance the performance of LLMs within the biomedical domain. Our future work will focus on further enhancing Taiyi's task capabilities, interpretability, and security for biomedical applications.

### Ethical Considerations

This research demonstrates the potential of LLMs in the biomedical domain. The transition from using Taiyi LLM for doctor-patient dialogue and medical report generation to practice applications in medical services will require abundant additional research to ensure the safety of this technology. In addition, strict expert evaluation for different medical scenarios must be considered in deployment to realize early diagnosis error discovery. Also, noteworthy aspects including biases and security vulnerabilities inherited from foundation models.

## FUNDING

This research was supported by the National Natural Science Foundation of China (No. 62302076) and the Fundamental Research Funds for the Central Universities [No. DUT23RC(3)014].

## AUTHOR CONTRIBUTIONS

Conception and design: LL, ZY, HL. Data collection and processing: LL, YZ, PC, WrF, YQ, DP, JL, HL, WdF, ST, YL. Model training and evaluation: JN, ZW, LL, ZD, QH, GX. Analysis and interpretation: LL, JN, YZ, YS. Drafting the manuscript: LL, JN, YZ, PC, ZD, WrF, QH. Revising the drafted manuscript: LL, JW, YS, ZY, HL. All authors approved the submitted version.

## SUPPLEMENTARY MATERIAL

Supplementary material is available at Journal of the American Medical Informatics Association online.

## CONFLICT OF INTEREST STATEMENT

None declared.

## DATA AVAILABILITY

The benchmark datasets that support the findings of this study are available from the official websites of natural language processing challenges with Data Use Agreements. The data information and model weights of Taiyi are available at https://github.com/DUTIR-BioNLP/Taiyi-LLM.

# Taiyi: A Bilingual Fine-Tuned Large Language Model for Diverse Biomedical Tasks

## Supplementary Material

Table S1. Summarization of existing LLMs in the biomedical domain

| Model Name | Base | Language | Training method | SFT data |
|---|---|---|---|---|
| GatorTron [1] | Transformer | En | PT+SFT | BioNLP |
| Med-PaLM [2] | PaLM | En | SFT | QA |
| ChatDoctor [3] | LLaMA | En | SFT | QA and Chat |
| MedAlpaca [4] | LLaMA | En | SFT | QA and Chat |
| PMC-LLaMA [5] | LLaMA | En | CPT+SFT | QA and Chat |
| BenTsao [6] | ChatGLM | Zh | SFT | QA and Chat |
| DoctorGLM [7] | ChatGLM | Zh | SFT | QA and Chat |
| HuatuoGPT [8] | BLOOMZ | Zh | SFT+RL | QA and Chat |
| ClinicalGPT [9] | BLOOM | Zh | SFT+RL | QA and Chat |
| DISC-MedLLM[10] | Baichuan | Zh | SFT | QA and Chat |
| Zhongjing [11] | Ziya-LLaMA | Zh | CPT+SFT+RL | BioNLP |
| ChiMed-GPT [12] | Ziya-LLaMA | Zh | CPT+SFT+RL | QA, Chat |

PT denotes pre-training from scratch; CPT denotes continued pre-training; SFT denotes supervised fine-tuning; RL denotes reinforcement learning.

Table S2. Examples of the instruction data

| Task: NER |
|---|
| **Ex1(zh):**<br>**Input:**<br>从下面文本中识别出指定的实体类型：<br>对儿童 SARST 细胞亚群的研究表明，与成人 SARS 相比，儿童细胞下降不明显，证明上述推测成立。<br>实体类型：微生物类，疾病，药物，医疗程序，医疗设备，临床表现，科室，身体，医学检验项目<br>**Output:**<br>疾病：成人 SARS<br>临床表现：细胞下降<br>身体：SARST 细胞亚群；细胞 |
| **Ex2(en):**<br>**Input:**<br>Identify Chemical, Disease entities from the text: "Acute changes of blood ammonia may predict short-term adverse effects of valproic acid. Valproic acid (VPA) was given to 24 epileptic patients who were already being treated with other antiepileptic drugs. A standardized loading dose of VPA was administered, and venous blood was sampled at 0, 1, 2, 3, and 4 hours. Ammonia (NH3) was higher in patients who, during continuous therapy, complained of drowsiness (7 patients) than in those who were symptom-free (17 patients), although VPA plasma levels were similar in both groups. By measuring VPA-induced changes of blood NH3 content, it may be possible to identify patients at higher risk of obtundation when VPA is given chronically.<br>**Output:**<br>Chemical: valproic acid; Ammonia; NH3; ammonia; VPA; Valproic acid |
| Disease: epileptic; drowsiness |
| **Task: RE** |
| **Ex1(zh)** <br>**Input:** <br>实体关系三元组抽取，以"(头实体，尾实体，关系类型)"格式输出： <br>13-三体综合征@30%～60%患儿有泌尿系畸形，可见多囊肾、肾盂积水、双肾及双输尿管。关系类型标签：并发症，发病率，预后状况，鉴别诊断，同义词，相关（转化），影像学检查，外侵部位，死亡率 <br>**Output:** <br>(13-三体综合征, 泌尿系畸形, 并发症); (13-三体综合征, 双肾, 并发症); (13-三体综合征, 多囊肾, 并发症); (13-三体综合征, 双输尿管, 并发症); (13-三体综合征, 肾盂积水, 并发症); (泌尿系畸形, 30%～60%, 发病率) |
| **Ex2(en)** <br>**Input:** <br>Output the chemical-induced disease relations in the following text: <br>Phenobarbital-induced dyskinesia in a neurologically-impaired child. A 2-year-old child with known neurologic impairment developed a dyskinesia soon after starting phenobarbital therapy for seizures. Known causes of movement disorders were eliminated after evaluation. On repeat challenge with phenobarbital, the dyskinesia recurred. Phenobarbital should be added to the list of anticonvulsant drugs that can cause movement disorders. <br>**Output**: <br>chemical-induced disease relations: [Phenobarbital, dyskinesia]; [phenobarbital, dyskinesia] |
| **Task: TC** |
| **Ex1(zh)** <br>**Input:** <br>将下面文本分类到指定的类别中："9. 曾接受过用于治疗癌症的免疫细胞治疗方法，包括 CIK、DC、DC-CIK、LAK 细胞等；" <br>类别标签：伦理审查，疾病，吸烟状况，预期寿命，依存性，肿瘤进展，受体状态，过敏耐受，实验室检查，年龄，性别，教育情况，研究者决定，健康群体，知情同意，酒精使用，体征(医生检测），口腔相关，药物，参与其它试验，器官组织状态，风险评估，锻炼，设备，护理，成瘾行为，读写能力，性取向，症状(患者感受)，献血，病例来源，数据可及性，特殊病人特征，怀孕相关，睡眠，治疗或手术，能力，饮食，残疾群体，种族，含有多类别的语句，居住情况，诊断，疾病分期 <br>**Output:** <br>上述文本被分类为：治疗或手术 |
| **Ex2(en)** <br>**Input:** <br>Classify the following text into the specified text label: "Influenza activity during the outbreak of coronavirus disease 2019 in Chinese mainland. Since coronavirus disease 2019 (COVID-19) might circulate in the following seasons, it is essential to understand how COVID-19 influences other respiratory diseases, especially influenza. In this study, we analyzed the influenza activity from mid-November 2019 to March 2020 in Chinese mainland and found that the influenza season ended much earlier than previous seasons for all subtypes and lineages, which may have resulted from the circulation of COVID-19 and measures such as travel control and personal protection. These findings provide rudimentary knowledge of the co-circulation patterns of the two types of viruses." <br>Text Labels: Case Report, Prevention, Transmission, Diagnosis, Mechanism, Treatment, Epidemic Forecasting <br>**Output:** <br>Result: Prevention |
| **Taks: MT** |
| **Ex1(zh)** <br>**Input:** <br>将下面文本翻译成英语："死亡的风险比(手术 vs.非手术)为 1.29(95%CI,0.97~1.72;P=0.08),相当于中位总生存期分别为 50.6 个月和 64.7 个月. <br>**Output:** <br>The hazard ratio for death ( surgery vs. no surgery ) was 1.29 ( 95 % confidence interval [ ci ], 0.97 to 1.72 ; p = 0.08 ), which corresponded to a median overall survival of 50.6 months and 64.7 months, respectively. |
| **Ex2(en)** <br>**Input:** <br>Machine Translation from English to Chinese: "Unfortunately, recurrent disease develops in more than 80 % of women." <br>**Output:** <br>但遗憾的是,有 80%患者的卵巢癌会复发. |

**Table S3**. Statistics of our evaluation test sets

| Task | Data | Lan | Size | Metrics |
|---|---|---|---|---|
| NER | BC5CDR-chem | En | 500 abstracts | Micro-F1 |
| | BC5CDR-dise | En | 500 abstracts | |
| | CHEMDERN | En | 3,000 abstracts | |
| | NCBI-Disease | En | 100 abstracts | |
| | BioRED | En | 100 abstracts | |
| | CMeEE-dev | Zh | 5,000 sentences | |
| RE | BC5CDR | en | 500 abstracts | Micro-F1 |
| | CMeIE-dev | Zh | 3,585 sentences | |
| TC | BC7LitCovid | En | 6,239 abstracts | Micro-F1 |
| | HOC | EN | 3,547 sentences | |
| | KUAKE_QIC-dev | Zh | 1,955 sentences | |
| QA-mc | PubMedQA | En | 500 questions | Accuracy |
| | MedQA-USMLE | En | 1,273 questions | |
| | MedQA-MCMLE | Zh | 3,426 questions | |

**Examples of Taiyi output on the supplementary biomedical NLP tasks**

In addition to the four primary evaluation tasks, the capabilities of the Taiyi model were also shown through examples (in Table S4) on several supplementary biomedical NLP tasks, such as **Medical Report Generation (MRG)** aims to automatically summarize information during the diagnostic process. The task requires the language model to generate medical reports in a specified format based on patient self-reports and doctor-patient dialogues. **Biomedical Event Extraction (EE)** is to extract structured biomedical events from unstructured biomedical literature, including mining the described biological entities in the literature, as well as the fine-grained relationships between biomedical events. **Biomedical Machine Translation (MT)** converts biomedical texts from one natural language to another. This task can assist in translating various medical questions and recommendations, disease names, symptoms, and other information between English and Chinese. **Biomedical Title Generation (TG)** aims to generate informative article titles for input biomedical text paragraphs with summarization. **Biomedical Text Semantic Similarity (SS)** aims to determine the degree of association or similarity between two medical texts. This task typically involves text matching, similarity measurement, and semantic understanding and can be used in various applications such as medical information retrieval and document matching. **Biomedical Question Answering and Chat (CHAT)** need to understand questions posed in natural language and provide accurate answers. Different from the simple QA, the dialogue task is capable of simulating multi-round natural language conversations.

**Table S4**. Example responses generated by Taiyi on the supplementary BioNLP tasks

**Task: Chat (en)**

**Input (1)**:
Do you know colchicine?
**Taiyi**:
Yes, Colchicine is a medication used to treat gout, which is a type of arthritis caused by high levels of uric acid in the blood. It works by reducing the production of uric acid in the body and also by relieving the pain and inflammation caused by the uric acid crystals in the joints.
**Input (2)**:
Does it have any potential side effects?
**Taiyi**:
Yes, Colchicine have potential side effects, including gastrointestinal issues such as nausea, vomiting, diarrhea, and abdominal pain. It can also cause allergic reactions such as rash, itching, hives, or difficulty breathing. In some cases, it can lead to serious complications such as anaphylaxis or kidney damage. Therefore, it is important to monitor patients closely while taking Colchicine and report any adverse effects immediately to the healthcare provider.

**Task: Medical Report Generation (zh)**

**Input:**
问诊对话历史：
患者：小孩受凉了，流清鼻涕，咳嗽，应该是风寒咳嗽，去药店买哪种药好呢
医生：你好，宝宝咳嗽，流涕比较常见，西医角度上呼吸道感染可能性大，中医上叫做风寒咳嗽，请问宝宝除了咳嗽有没有其他不适症状呢？例如发热等，请详细描述一下，我好帮你诊治分析病情
患者：精神状态好，也没有发热，就是喉咙有一点痛，咳嗽
医生：先帮你分析一下病情，宝宝受凉之后免疫力降低，就会被细菌或病毒侵袭体内，气道分泌物增多，支气管平滑肌痉挛，咳嗽，咳痰，咽通。
医生：目前没有发热，宝宝病情不重，不用过分紧张的。
医生：我帮推荐治疗方法
医生：宝宝目前多大了？有没有再医院看过？做过化验检查
患者：嗯
患者：7岁，没去医院，做过很多检查，平常就是爱咳嗽，喉咙发炎
患者：医生说，扁桃体偏大
医生：近期这次有没有去医院看过？做过检查
医生：如果宝宝没有其他不适？可以口服氨溴索，桔贝合剂效果好
医生：另外如果条件允许，可以做做雾化吸入治疗直接作用与支气管粘膜，效果更直接
患者：不用做雾化吧，吃点药就行了
医生：也可以先吃药
患者：近期没有去过
医生：你们这次没有去医院看过？
患者：要吃消炎的吗
患者：没
患者：要吃消炎药吗
医生：你好，可以先不吃的
患者：那家里有蒲地蓝，可以吃吗
患者：口服液
患者：喉咙痛要吃吗
医生：先治疗看看，可以吃的，假如宝宝出现发热或咳嗽加重，医院就诊，复查血常规和胸片，那个时候再考虑加抗生素
患者：另外买个止咳的，行吗
医生：我们的观点是宝宝小，尽量少吃消炎药，可以先吃那几个药三天看看效果
患者：嗯谢谢
根据上述对话，给出诊疗报告
说明：诊疗报告分为主诉，现病史，辅助检查，既往史，诊断，建议这六个章节。

**Taiyi:**
上述问诊对话的诊疗报告如下：
主诉：咳嗽咳痰伴鼻流涕。 现病史：患儿无明显诱因下出现咳嗽咳痰伴鼻流涕情况，无恶心呕吐，无其他不适，无其他疾病史，无药物过敏史。 辅助检查：无。 既往史：既往体健。 诊断：上呼吸道感染。 建议：口服氨溴索，桔贝合剂，若条件允许，雾化吸入治疗。

**Taks: Event Extraction (zh)**

**Input**:

找出指定的临床发现事件属性：
因患者需期末考试，故予以口服"雷贝拉唑钠肠溶片"治疗，现腹痛情况明显好转。
事件抽取说明：临床发现事件由主体词，发生状态，描述词和解剖部位组成
**Taiyi**:
上述句子中的临床发现事件如下：
主体词：疼痛；发生状态：；描述词：明显好转；解剖部位：腹

### Taks: Event Extraction (en)

**Input**:
Input text: "Contaminated drinking water is responsible for causing diarrheal diseases that kill millions of people a year.
Even Types: Treatment of disease, Cause of disease
Role Types: Cause, Theme
Please extract events from the input text.
**Taiyi**:
Cause of disease: (Trigger: responsible, Theme: diarrheal diseases, Cause: Contaminated drinking water]

### Task: Machine Translation (zh-en)

**Input**:
将下面文本翻译成英文：
光动力疗法（PDT）作为一种新兴的肿瘤治疗手段，因其不良反应较少、靶向性好、可重复治疗等优点，已广泛应用于临床多种肿瘤的治疗。相比于手术、化疗及放疗等传统治疗策略，光动力疗法不仅可杀伤原位肿瘤，还可通过激活机体的免疫效应对转移瘤发挥抑制作用。然而，PDT 诱导免疫效应的高低受多种因素影响，包括光敏剂在细胞内的定位和剂量、光参数、肿瘤内的氧浓度、免疫功能的完整性等。本文针对 PDT 介导抗肿瘤免疫效应的相关机制，以及 PDT 免疫效应的主要影响因素进行综述，以探讨 PDT 用于肿瘤治疗的未来发展方向。
**Taiyi**:
As a new tumor treatment, photodynamic therapy (PDT) has been widely used in clinical practice due to its advantages of fewer side effects, high target specificity, and the possibility of repeated treatment. Compared with traditional surgical, chemotherapy, and radiotherapy, PDT can not only kill the primary tumor, but also inhibit the metastatic tumor by activating the body's immune effect. However, the high or low of PDT-induced immune effect is affected by many factors, including the localization and dosage of photosensitizers, light parameters, oxygen concentration in tumor, and the integrity of immune function. This paper reviews the related mechanisms of PDT-mediated anti-tumor immune effect and the main influencing factors, and discusses the future development direction of PDT in tumor treatment.

### Task: Machine Translation (en-zh)

**Input**:
Translate the following text into Chinese:
Metabolic-associated fatty liver disease (MAFLD), formerly known as non-alcoholic fatty liver disease (NAFLD), is one of the most popular chronic liver diseases in the world. There are many explorations on the epidemiology and risk factors of the disease published in recent years. With the increasing burden of MAFLD on medical care and social economy, drug treatment of MAFLD has become an urgent research hotspot. Based on this, the current epidemiological status, risk factors, diagnostic criteria and treatment of MAFLD were summarized in this review in order to enlighten the clinical practice and scientific research of MAFLD.
**Taiyi**:
代谢综合征相关性脂肪肝（MAFLD）是世界范围内最为流行的慢性肝病之一。近年来，该病的流行病学及危险因素等方面的研究颇多。随着 MAFLD 在医疗保健及社会经济方面所造成的负担日益加重，该病的药物治疗已成为研究热点。本文对该病的流行病学现状、危险因素、诊断标准及治疗等方面进行了综述，以期为临床实践及 MAFLD 的科学研究提供参考。

### Task: Title Generation (zh)

**Input**:
请给下面摘要起标题：
气管食管瘘是指气管或支气管与食管之间的病理性瘘道，包括气管-食管瘘和支气管-食管瘘，临床以气管-食管瘘较多见。气管食管瘘致病原因较多，可引起严重的并发症，是对患者生活质量影响显著、治疗困难和病死率较高的疾病。气管食管瘘目前治疗方式较多，但多数疗效欠佳，对新兴治疗手段的需求迫切。胸腹部 X 线摄影检出鼻胃管滞留是气管食管瘘诊断的金标准，其主要治疗方法包括外科手术治疗、支架置入、局部生物胶水封闭、干细胞治疗等。本文综述近年气管食管瘘诊断与治疗的主要研究进展，旨在为该病的临床诊治提供参考。
**Taiyi**:
气管食管瘘的诊断与治疗进展

### Task: Title Generation (zh)

**Input**:
Output a title for the following abstract:
The incidence of diabetes mellitus has been increasing, prompting the search for non-invasive diagnostic methods. Although current methods exist, these have certain limitations, such as low reliability and accuracy, difficulty in individual patient adjustment, and discomfort during use. This paper presents a novel approach for diagnosing diabetes using high-frequency ultrasound (HFU) and a convolutional neural network (CNN). This method is based on the observation that glucose in red blood cells (RBCs) forms glycated hemoglobin (HbA1c) and accumulates on its surface. The study incubated RBCs with different glucose concentrations, collected acoustic reflection signals from them using a custom-designed 90-MHz transducer, and analyzed the signals using a CNN. The CNN was applied to the frequency spectra and spectrograms of the signal to identify correlations between changes in RBC properties owing to glucose concentration and signal features. The results confirmed the efficacy of the CNN-based approach with a classification accuracy of 0.98. This non-invasive diagnostic technology using HFU and CNN holds promise for in vivo diagnosis without the need for blood collection.
**Taiyi**:
Application of high-frequency ultrasound and convolutional neural network in non-invasive diagnosis of diabetes

### Task: Text Classification (zh)

**Input**:
现有以下文本：
治皮肤病费用大概多少？
请将上述文本分类至指定类别中：医疗费用，后果表述，指标解读，病情诊断，就医建议，疾病描述，其他，治疗方案，病因分析，功效作用，注意事项。
**Taiyi**:
上述文本被分类为：医疗费用

### Task: Text Classification (en)

**Input**:
Document triage: "Will my mask from sherwin williams paint store with filters protect me from corona virus along with paint fumes?"
Labels: patient, doctor
**Taiyi**:
Result: patient

### Task: Text Semantic Similarity (zh)

**Input**:
语句 1：乙肝小二阳会转成小三阳吗？
语句 2：乙肝小三阳会不会转成肝硬化、肝癌？
请从下面选项中评估这段文本的语义相似度：语义不相同，语义相同
**Taiyi**:
语义不相同

### Task: Text Semantic Similarity (en)

**Input**:
1. How can someone's happiness level affect someone's health?
2. Can staying happy improve my health? What specific steps should I take?
Assess the semantic similarity of the text pairs based on the following labels: dissimilar, similar
**Taiyi**:
similar

## Case study

To further investigate the benefits of the in-domain instruction fine-tuning, we qualitatively analyzed model outputs on various BioNLP tasks and compared Taiyi to Qwen7B-Chat, a published LLM fine-tuned using general domain instruction data on the same base model Qwen7B. Table S5 shows some examples.

**Table S5**. Some cases of generated text on the BioNLP tasks

### Taks: NER (zh)

**Input**:
从以下文本中识别出指定的实体类型：

两肺听诊可闻及干湿性啰音，胸部 X 线片两肺有斑片状模糊阴影，尤以双下肺明显，严重病例可合并胸腔积液或脓胸。
实体类型：疾病，医疗程序
**Taiyi**：
疾病：胸腔积液；脓胸
医疗程序：两肺听诊；胸部 X 线片
**Qwen-Chat**：
病：肺炎
医疗程序：听诊、胸部 X 线检查

**Task: NER (en)**

**Input**:
Identify Chemical, Disease entities in the following text:
Phenobarbital-induced dyskinesia in a neurologically-impaired child. A 2-year-old child with known neurologic impairment developed a dyskinesia soon after starting phenobarbital therapy for seizures. Known causes of movement disorders were eliminated after evaluation. On repeat challenge with phenobarbital, the dyskinesia recurred. Phenobarbital should be added to the list of anticonvulsant drugs that can cause movement disorders.
**Taiyi**:
Chemical: Phenobarbital; phenobarbital
Disease: dyskinesia; neurologically-impaired; neurologic impairment; seizures; movement disorders
**Qwen-Chat**:
Chemical: Phenobarbital, Anticonvulsant drugs
Disease entities: Neurological impairment, Seizures, Movement disorders

**Task: RE (zh)**

**Input**:
出下面文本中具有药物治疗关系的实体对：
疟疾感染应当使用氯喹（首选）或羟氯喹治疗感染，共 3 次给药。患者应遵循医生的建议和药物使用说明，以确保药物的最大效力。
**Taiyi**：
具有药物治疗关系的实体对如下：[疟疾感染, 氯喹]; [疟疾感染, 羟氯喹]
**Qwen-Chat**：
该句包含以下药物实体对：
氯喹：药物名称
羟氯喹：药物名称

**Task: RE (en)**

**Input**:
Extract the Protein-Protein Interaction from the following text:
Such motifs are known to bind to profilins (PFNs), small proteins engaged in the control of actin dynamics.
**Taiyi**:
[actin, PFNs]; [actin, profilins]
**Qwen-Chat**:
The Protein-Protein Interaction mentioned in the given text is between "motifs" and "profilins" (PFNs).

From the results, Taiyi generates more detailed and accurate responses to biomedical questions compared to the general domain model Qwen-Chat. For the Chinese NER case, Qwen-Chat incorrectly extracts entities, while Taiyi precisely identifies disease and medical procedure entities. Similarly, Taiyi identifies more correct chemical entities than Qwen-Chat in the English NER case. In the final case of relation extraction, Taiyi properly extracts complex biomedical relations between proteins, diseases, and chemicals, which Qwen-Chat is unable to capture. It is difficult for the model that has not been fine-tuned with domain task instructions to understand the biomedical relation extraction instructions. These examples highlight the advantages of domain-specific instruction fine-tuning for enhancing Taiyi's performance on diverse BioNLP tasks compared to the model fine-tuned with general instructions. More examples on different BioNLP tasks can be found in the Supplementary Material.